\documentclass[11pt,a4paper]{article}
\usepackage[hyperref]{eacl2021}
\usepackage{xcolor, soul}
\usepackage{url}  
\usepackage{todonotes}
\usepackage{graphicx}
\usepackage{longtable}
\usepackage{epigraph}
\usepackage{times}
\usepackage{float}
\usepackage{pgfplots}
\usepackage{tikz}
\usepackage{subcaption}
\pgfplotsset{compat=1.15}
\usepackage{latexsym}

\usepackage{microtype}

\aclfinalcopy 
\def\aclpaperid{***} 

\title{I Beg to Differ:\\ A study of constructive disagreement in online conversations}
\author{Christine De Kock \& \textbf{Andreas Vlachos}\\
  University of Cambridge\\
  Department of Computer Science and Technology\\
  \texttt{\{cd700,av308\}@cam.ac.uk}}
\date{}

\begin{document}
\maketitle
\begin{abstract}

Disagreements are pervasive in human communication. In this paper we investigate what makes disagreement constructive. To this end, we construct WikiDisputes, a corpus of 7\,425 Wikipedia Talk page conversations that contain content disputes, and define the task of predicting whether disagreements will be escalated to mediation by a moderator. We evaluate feature-based models with linguistic markers from previous work, and demonstrate that their performance is improved by using features that capture changes in linguistic markers throughout the conversations, as opposed to averaged values. We develop a variety of neural models and show that taking into account the structure of the conversation improves predictive accuracy, exceeding that of feature-based models. 
We assess our best neural model in terms of both predictive accuracy and uncertainty by evaluating its behaviour when it is only exposed to the beginning of the conversation, finding that model accuracy improves and uncertainty reduces as models are exposed to more information. 

\end{abstract}

\section{Introduction}\label{sec:intro}
\begin{figure}[t!]
\centering
\setlength\fboxsep{0pt}
\setlength\fboxrule{0.25pt}
\includegraphics[width=0.47\textwidth]{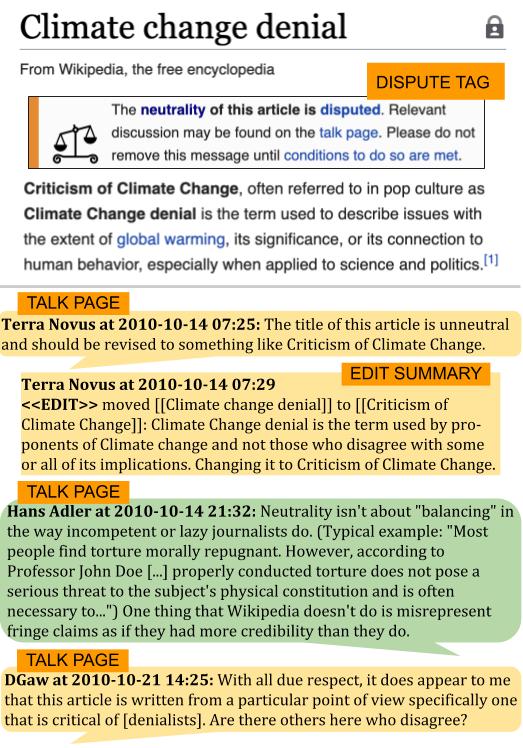}
\caption{An example of a dispute on Wikipedia, showing the dispute tag on the article, talk page posts, and summaries of edits occurring during the discussion.}
\label{fig:example_p1}
\end{figure}

Disagreements online are a familiar occurrence for any internet user. While they are often perceived as a negative phenomenon, disagreements can be useful; as is illustrated in Figure~\ref{fig:example_p1}, disagreements can lead to an improved understanding of a topic by introducing and evaluating different perspectives. Research on online disagreements has focused on mostly negative aspects such as trolling \citep{cheng2017anyone}, hate speech \citep{waseem-hovy:2016:N16-2}, harassment \citep{yin2009detection} and personal attacks \citep{wulczyn2017ex}. Recent works by \citet{zhang2018conversations} and \citet{chang-danescu-niculescu-mizil-2019-trouble} study conversations on Wikipedia talk pages that start out as civil but derail into personal attacks, using  linguistic markers and deep learning approaches, respectively.

An alternative approach is to study good faith disagreement through debates on online platforms such as ChangeMyView \citep{tan2016winning}, debate.org \citep{durmus-cardie-2019-corpus} and Kialo \citep{boschi2019having}, or formal Oxford-style debates \citep{zhang-etal-2016-conversational}. While this has benefits, such as readily available annotation of stances and indication of winning and losing sides, 
it does not mirror the way people naturally converse. For example, in formal debates, temporal and structural constraints are imposed. Moreover, in Oxford-style debates, participants are not motivated to come to a consensus but rather to persuade an audience of a predefined stance
\citep{zhang-etal-2016-conversational}. Finally, the task of detecting disagreement in conversations has been studied by a number of authors, e.g.\ \citet{wang-cardie-2014-piece} and \citet{rosenthal-mckeown-2015-couldnt}, without attempting to analyze it further. 

We are interested instead in \textit{constructive disagreement in an uncoerced setting}. To this end, 
we present \texttt{WikiDisputes}\footnote{\url{github.com/christinedekock11/wikidisputes}}; a corpus of 7\,425 disagreements (totalling 99\,907 utterances) mined from Wikipedia Talk pages. To construct it, we map dispute tags in the platform's edit history (shown in Figure~\ref{fig:example_p1}) to conversations in WikiConv \citep{hua-etal-2018-wikiconv} to locate conversations that relate to content disputes. Observing that 
conversations are often conducted in the Talk pages and the edit summaries simultaneously, as illustrated in Figure~\ref{fig:example_p1}, we augment the WikiConv conversations with edit summaries that occur concurrently. To investigate the factors that make a disagreement constructive, we define the task of predicting whether a dispute is eventually considered resolved by its participants, or it was escalated by them to mediation by a moderator, and therefore considered unconstructive. 

We evaluate both feature-based and neural models for this task. Our findings indicate that balancing hedging and certainty plays an important role in constructive disagreements. We find that incorporating edit summaries (which have been ignored in previous work on Wikipedia discussions) improves model performance on predicting escalation. We further find that including information about the conversation structure aids model performance. We observe this in two ways. Firstly, we include gradient features, which capture changes in linguistic markers throughout the conversations as opposed to averaged values. Secondly, we experiment with adding sequential and hierarchical conversation structure to our neural models, finding that a Hierarchical Attention Network \citep{yang2016hierarchical} provides the best performance on our task.

We further evaluate this model in terms of its predictive accuracy when exposed to the beginnings of the conversation as opposed to completed conversations, as well as its uncertainty (more details in Section \ref{sec:uncertainty}). Our results indicate that model performance is reduced from a PR-AUC of 0.29 to 0.19 when exposed to only the first half of the conversation. However, this reduced model still outperforms toxicity and sentiment models predicting on full conversations. Model uncertainty decreases roughly linearly as the model is exposed to more information. We conduct a qualitative analysis of the points in the conversation where the model changes its prediction on constructiveness, finding that some markers from our feature-based models also seem to have been picked up by the neural model.  

\section{Dispute resolution on Wikipedia}
\begin{figure*}
    \centering
    \includegraphics[width=\textwidth]{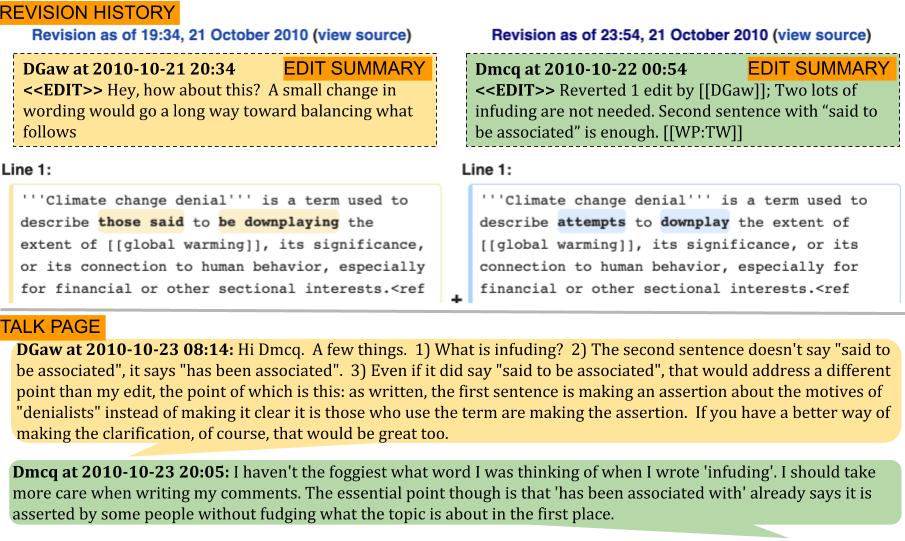}
    \caption{An illustration of the relationship between the revision history and Talk page conversations, using later utterances from the 
    dispute in Figure \ref{fig:example_p1}. Two consecutive edits and their summaries are shown side by side.}
    \label{fig:example_p2}
\end{figure*}

\label{sec:disp_res}
Wikipedia relies heavily on collaboration and discussion by editors to maintain content standards, using the platform's Talk pages \citep{wiki_dispres_policy}. Wikipedia contributors add ``dispute'' templates to articles \citep{wiki_disp_tag} referring to a content accuracy dispute or to a violation of the platform's neutral point of view (NPOV) policy \citep{wiki_npov_tag}. Adding such a template to the article creates a dispute tag on the article as shown in Figure~\ref{fig:example_p1}. Variations of these tags allow contributors to flag specific sections or phrases which relate to the issue in question, or to add details of the dispute. 

All edits to Wikipedia, including the aforementioned dispute tags, are retained in the site's edit history. Revision metadata contains a description of the edit (hereafter referred to as the edit summary). As observed in Figure~\ref{fig:example_p1}, when such edits occur while a conversation is underway, editors use the ``edit summaries'' to clarify their intentions in the context of the conversation.

The Wikipedia dispute resolution policy stipulates that if contributors are unable to reach a consensus through Talk page discussion, they can request mediation by a community volunteer. A prerequisite for this escalation step is proof of recent, extensive discussion on a Talk page\footnote{\url{https://en.wikipedia.org/wiki/Wikipedia:Dispute_resolution_noticeboard}}. 

\section{WikiDisputes}\label{sec:wikidisputes}
Our dataset consists of three facets: disagreements on Talk pages, edit summaries, and escalation labels. 
Wikipedia informs users that all contributions on article or Talk pages are published publicly, and provides channels for users to permanently remove personal information that they have shared accidentally or otherwise \citep{wiki_personal_security}, thus we are able to release this data to the community. In the remainder of this section, we detail the process for obtaining the conversations, the edit summaries and the labels. 

\subsection{Finding disagreements}\label{sec:finding_disagreements}
We use the Wikipedia revision history dump\footnote{\url{https://dumps.wikimedia.org/enwiki/}} from 1 July 2020 to find content disputes. We locate the addition of dispute templates in the article history using regular expressions that account for variations in the free text dispute templates. Using the revisions where dispute tags are added, we can also determine which user logged the dispute, the timestamp, and the article and section in dispute.

To find the related WikiConv conversation for each dispute tag, we select all conversations on the relevant article's Talk page in which the user who logged the dispute was involved. Of these candidates, the conversation closest in time to the timestamp of the dispute tag addition is selected. We filter the conversations extracted above using a number of heuristics in the pursuit of high quality samples of constructive disagreements. To ensure conversations of adequate length, we set a minimum conversation length of five utterances and 250 tokens. Based on an inspection of 100 conversations, we remove conversations of more than 50 utterances, as such conversations are often flame wars. We further include only conversations in which more than one user participated. Using this methodology, we obtain 7\,425 content disputes.

Our manual inspection further indicates that the usage of the dispute tag is not entirely consistent across the platform; editors sometimes use the tag to report that they dispute some aspect of the text before discussing it, rather than to flag a currently occurring dispute in the Talk page. However, the former frequently results in a disagreement. We found that 88 out of 100 inspected examples were correctly identified disagreements, which we believe is an acceptably low error rate.

\subsection{Collecting edit summaries}\label{sec:collecting_edit_summaries}
We show in Figure~\ref{fig:example_p2} two further posts from the climate dispute (continued from Figure~\ref{fig:example_p1}), along with two edits that occurred during the course of the conversation. We note here that including the edit summaries as part of the conversation is important for understanding the conversation; without this, the reference to ``infuding'' in this example would be incomprehensible.
We thus include all edit summaries written by users involved in the conversation, which are timestamped between the first and last utterance in the conversation. 

\subsection{Escalation tags}\label{sec:escalation_tags}
For the task of inferring whether a dispute was escalated to mediation (and therefore Talk page discussion was unsuccessful), we scrape mediated cases from the Dispute Resolution Noticeboard archives and align them with WikiConv conversations.

Of the 2 520 archived mediation processes, 149 were closed as failures and 237 as successes, with the remaining 2 134 receiving ``General closures''. The last label is frequently assigned due to insufficient Talk page discussion. We only include accepted mediations in our dataset and thus ignore ``General closures'', resulting in 386 mediations.

We record the location of the disagreement to which the mediation relates, the usernames of participants and the timestamp. We use a similar alignment procedure to that described in Section \ref{sec:finding_disagreements} to find the relevant WikiConv conversations; except that we include the usernames of all participants listed in the mediation. During alignment, a simple match is recorded when there is exactly one conversation on the article Talk page involving all listed users (118 cases out of 386). 

There are 111 cases in which no full matches occur; this sometimes happens when a participant changes their username or is participating anonymously. We manually inspect potential matches in this case. A further 67 cases have multiple matches; this happens when a participant changes their username, or is participating anonymously, or the Talk page has been archived; in this case we try to find the conversation most related to the mediation summary.

We use only the utterances which are timestamped from before the conversation was escalated and apply the same filters outlined in Section \ref{sec:finding_disagreements}. Using this methodology, we are able to find 201 disagreements that are escalated to mediation. The complementary class label (not escalated) is assigned to the disagreements extracted using the dispute tags, for which no mediation was found. 

Our classification task uses escalation as a proxy label for cases where Talk page discussion was not constructive. This relies on the assumption that non-escalated disagreements are more constructive than escalated disagreements, even though disagreements that were not escalated are not guaranteed to be successfully resolved; participants may simply not be aware of the escalation procedure, or not be willing to go through the extra effort of participating in mediation. To validate this, we annotated 35 samples from the dataset as to whether they represented constructive disagreements, and our annotations agreed with the escalation labels in 25 of these.
Combined with the low estimates of antisocial behaviour on the site \citep{wulczyn2017ex}, we draw the inference that the non-escalated class contains more constructive disagreements than the escalated ones. Throughout the remainder of this paper, we use these terms interchangeably. 

\section{Modelling constructive disagreement}\label{sec:method}

\subsection{Feature-based models} 

We develop our feature-based models using feature sets suggested in previous research on tasks related to conversational analysis.
These include:
\begin{itemize}
\itemsep0em
    \item \textbf{Politeness}: The politeness strategies from \citet{zhang2018conversations} as implemented in Convokit \citep{chang-etal-2020-convokit}, which capture greetings, apologies, directness, and saying ``please'', etc.
    \item \textbf{Collaboration}: Conversation markers in collaborative discussions \citep{niculae-danescu-niculescu-mizil-2016-conversational}, which capture the introduction and adoption of ideas, uncertainty and confidence terms, pronoun usage, and linguistic style accommodation. 
    \item \textbf{Toxicity}: Toxicity and severe toxicity scores as estimated by the Perspective API \cite{wulczyn2017ex} and included in WikiConv \citep{hua-etal-2018-wikiconv}.
    \item \textbf{Sentiment}: Positive and negative sentiment word counts, as per the lexicon of \citet{liu2005opinion} and implemented in Convokit \citep{chang-etal-2020-convokit}.
\end{itemize}

For each feature, we calculate the average value throughout the conversation, as well as the gradient of a straight line fit of the feature value throughout the conversation. Our intuition for including this variation is that the outcome of a disagreement is likely to be influenced by a change in the language usage through the course of a conversation, which is not reflected by the mean. For instance, a disagreement might start out very politely and end impolitely, or the other way around, and would have the same mean. Logistic regression is used to infer linear relationships between the linguistic features and disagreement outcomes. 

\subsection{Neural models}\label{sec:models}
Dialogue structure has been difficult to capture in feature based models due to sparsity. 
For this reason, we implement a number of neural models with increasing capacities for modelling conversation structure to assess its importance. 

\paragraph{Averaged embeddings} Our simplest variant averages the GloVe embeddings \citep{pennington-etal-2014-glove} of all words in a conversation, and uses a fully connected layer for classification. It ignores both utterance hierarchy
and word ordering and can be seen as a bag-of-word-embeddings approach.

\paragraph{LSTM} Instead of averaging the GloVe embeddings, we use a bidirectional LSTM-based model \citep{hochreiter1997long} to process the sequence of words in the conversation; ignoring utterance hierarchy but preserving word order.

\paragraph{HAN} We use a Hierarchical Attention Network (HAN) \citep{yang2016hierarchical} to model both word order and utterance hierarchy. HANs have recently been used to predict emotion in conversations \citep{ma2020han} and are useful for capturing the structure of texts by building up utterance embeddings from word embeddings, and then constructing a conversation vector from these utterance embeddings. Given a sequence of words in an utterance, the words are first mapped to vectors through an embedding matrix (GloVe embeddings, in our case). A bidirectional LSTM layer \citep{hochreiter1997long} is used to process the sequence of embeddings and calculate an embedding for each word in the context of the utterance. An attention mechanism \citep{bahdanau2015neural} is applied to these embeddings to find an utterance embedding. A similar procedure is followed to build up a conversation vector based on the utterances embeddings. This vector is then used to perform the classification. 
To incorporate edit summaries in a conversation, we prepend each edit summary with a special token ($<$EDIT$>$) to indicate its origin.

\section{Experimental setup}
An initial data analysis indicated that 
escalated disagreements have a median length of 16 utterances, compared to 11 for their non-escalated counterparts. While this may be an informative feature for classification, we are primarily interested in the effects of language on the conversation outcome, and therefore choose to control for this effect. We use \textit{matching}, a technique developed for causal inference in observational studies \citep{rubin2007design}, also employed by \citet{zhang2018conversations} in the context of conversational modelling. We pair each escalated disagreement with a non-escalated disagreement of the same length in utterances, bearing in mind that the dataset is heavily imbalanced, with 7\,425 non-escalated disagreements compared to 201 escalated disagreements. To retain an imbalance in the classes while conducting matching, we match every escalated disagreement with up to ten non-escalated disagreements (the actual number depending on availability) by randomly sampling without replacement from the non-escalated disagreements of the same length. Characteristics of the dataset after performing matching are shown in Table~\ref{tab:dataset_stats}, with both classes having a median of 16 utterances now. Escalated disagreements have one more participant than non-escalated disagreements on average, and utterances in the escalated disagreements class are slightly shorter. 

\begin{table}
    \small
    \centering
    \begin{tabular}{|l|l|l|}
    \hline
    & \textbf{Not escalated} & \textbf{Escalated} \\
    \hline
    \# Samples & 1994 &  216 \\
    \# Participants    &    2     &  3 \\
    \# Utterances    &    16    & 16 \\
    Tokens per utt. &    61      &   53 \\
    \hline
    \end{tabular}
    \caption{Dataset statistics for the task of predicting escalation. Where applicable, median values are used.}
    \label{tab:dataset_stats}
\end{table}

We split the dataset for training as indicated in Table~\ref{tab:escalation_dataset}. Due to the class imbalance we use the area under the precision-recall curve \citep{davis2006relationship} as metric for this task; however, for the sake of interpretability we also present the break-even F1 scores. We use a distribution-aware random class predictor as a random baseline. 

\paragraph{Hyperparameters} The logistic regression models are implemented in Scikit-Learn. We use a grid search to determine the best regularisation mode (L1 or L2) per model, evaluating C-values in [0.1,1,10,100]. The neural models are implemented in Keras, using Focal Loss \citep{lin2017focal} with Adam optimisation \citep{kingma2014adam} (learning rate=0.001) and Dropout ($p=0.3$). For the HAN, we use bidirectional LSTM layers with 128 nodes in both the utterance and conversation encoders. For the LSTM model, we use only one such a layer.

\section{Results}
Results are shown in Table~\ref{tab:escalation_results}. We note that generally, the neural network models perform better than the feature-based models. We discuss the PR-AUC scores of each model category below, noting that the scores from the F1 metric generally follow the same trends, but that PR-AUC is more robust in the face of imbalanced data. Furthermore, we investigate whether it is possible to predict the outcome from the beginning, how model uncertainty changes, or if there is often some inflection point in a conversation that changes the outcome.

\begin{table}[]
    \small
    \centering
    \begin{tabular}{|l|l|l|}
    \hline
    \textbf{Dataset} & \textbf{Escalated} & \textbf{Not escalated} \\
    \hline
       Train  &  125 & 1411 \\
       Test  &  46 & 284 \\
       Validation  & 30 & 299 \\
       \hline
    \end{tabular}
    \caption{Test set splits for predicting escalation.}
    \label{tab:escalation_dataset}
\end{table} 

\begin{table}[]
    \small
    \centering
    \begin{tabular}{|l|l|l|}
    \hline
        \textbf{Model} & \textbf{PR-AUC} & \textbf{F1}\\
        \hline
        \multicolumn{3}{|c|}{\textbf{Baselines}}\\
        \hline
        Random  & 0.121 & 0.128\\
        Bag-of-words &  0.213 & 0.239\\
        \hline
        \multicolumn{3}{|c|}{\textbf{Feature-based models}}\\
        \hline
        Toxicity & 0.140 & 0.125\\
        Sentiment & 0.150 & 0.055\\
        Politeness & 0.232 & 0.241\\
        \hspace{0.5cm} \textit{$+$ gradients} & 0.275 & 0.243\\
        Collaboration & 0.261 & 0.320\\
        \hspace{0.5cm}\textit{ $+$ gradients} & 0.269 & 0.302\\
        Politeness and collaboration & 0.255 & 0.256\\
        \hspace{0.5cm} \textit{$+$ gradients} & 0.281 &  0.289\\
        \hline
        \multicolumn{3}{|c|}{\textbf{Neural models}}\\
        \hline
        Averaged embeddings & 0.243 & 0.256\\
        LSTM & 0.263 & 0.194 \\
        HAN & 0.373 & 0.304 \\
        \hspace{0.5cm}\textit{$+$ edit summaries}& \textbf{0.400} & \textbf{0.333}\\
        \hline
    \end{tabular}
    \caption{Results of the escalation prediction task.}
    \label{tab:escalation_results}
\end{table}

\subsection{Feature-based models}\label{sec:results_features}
A number of our feature-based models outperforms the bag-of-words baseline, which in turn performs better than the random baseline. The toxicity and sentiment features do not perform better than the bag-of-words baseline, which indicates that these features by themselves do not capture constructiveness in disagreements. Toxicity scores were found by \citet{zhang-etal-2016-conversational} to be predictive of a conversation derailing, but in our case seemingly lead to the inference of spurious correlations. An explanation might be that comments which seem toxic out of context are in fact spirited discussion due to the degree of involvement of the participants; for instance, we have observed that contributors in some cases challenge others' credentials, which may seem rude in casual conversation, but can be useful in resolving a dispute. 

The best featuresets proposed in previous work are collaboration \citep{niculae-danescu-niculescu-mizil-2016-conversational}, followed closely by politeness \citep{zhang-etal-2016-conversational}. Adding the gradient features we proposed improves performance for both, indicating that not only the presence of a marker is important, but also how its usage throughout a conversation changes. The best feature-based model is a combination of politeness and collaboration features, with a PR-AUC of 0.281 ($P<0.05$, using a randomized permutation test). 

\begin{table}[]
    \small
    \centering
    \begin{tabular}{|p{3.7cm}|l|l|}
        \hline
        \textbf{Feature} & \textbf{Type} & \textbf{Coeff.} \\
        \hline
        2nd person pronouns, $\bar{x}$	& Both & +3.64 \\
        \# hedging terms, $\bar{x}$	& Both & -2.48	\\
        Greetings, $\bar{x}$ & Politeness &	+2.30 \\
        1st person pronouns, $\bar{x}$ & Both & +1.91 \\
        Greetings, $\nabla$ & Politeness& -1.81	\\
        Deference, $\bar{x}$ & Politeness& 	-1.44 \\
        3rd person pronouns, $\nabla$ & Collaboration &	-1.38 \\
        \# ideas adopted $+$ certainty, $\bar{x}$ & Collaboration&	-1.23 \\
        Use of ``by the way'',  $\bar{x}$& Politeness	& -1.04	\\
        Certainty,$\nabla$ & Collaboration&	-0.92 \\
        \hline
    \end{tabular}
    \caption{Ten features with the largest coefficients of the best feature-based model, which uses politeness and collaboration featuresets. Feature variations are the mean ($\bar{x}$) and gradient ($\nabla$). Positive weights are associated with the unconstructive class.}
    \label{tab:coef_constructiveness}
\end{table}

The ten features with the largest coefficients from the combined model are shown in Table~\ref{tab:coef_constructiveness} in descending order or magnitude, along with their directions. Positive coefficients are associated with the escalated class, which indicates that a disagreement was not constructive. 

Politeness markers such as deference and the use of ``by the way'' are found indicative of constructiveness, corroborating the findings of \citet{zhang2016conversational}. Greetings are associated with unconstructive disagreements on average, but an increase in greetings towards the end of a conversation is associated with constructiveness. An explanation of this might be that greetings can seem overly formal or indicative of tension early on, but if used later in a conversation, they indicate that new participants have entered the conversation or that more time is taken between replies. Indeed, longer gaps between replies (a feature in the collaboration featureset) are also associated with constructiveness. 

The use of first and second person pronouns (``I'' and ``you'') is associated with unconstructive disagreements, with the latter
corroborating the findings of
\citet{zhang2018conversations}.
This is also consistent with psychotherapy
research on disagreements in relationships (e.g.\ \citet{gottman1989marital}), which emphasises avoiding `you'-messages which might be perceived as blameful.

Hedging is associated with constructive disagreements, which corroborates the findings of \citet{zhang-etal-2016-conversational}. An intuitive understanding of this result is that using hedging terms shows that the speaker is more open to adjusting their opinion or compromising. The certainty gradient and the adoption of new ideas with certainty are also associated with constructiveness. \citet{niculae-danescu-niculescu-mizil-2016-conversational} found no significant correlation between either certainty or hedging and successful collaboration. We attribute the observed differences in feature associations to the fact that WikiDisputes are sourced from an uncoerced setting, where users feel more invested in the outcome of a disagreement and may need to balance the use of certainty and hedging to negotiate compromises. The setting of \citet{niculae-danescu-niculescu-mizil-2016-conversational} obligates volunteers to participate in a photo geolocation game, where it is unlikely that interlocutors are as invested in the task as collaborators working on a widely read encyclopaedia. 

Words associated with either class, obtained through the bag-of-words model, are shown in Appendix \ref{app:words}. An interesting observation from this list is that citing Wikipedia policy (which is indicated with WP) is associated with escalating conversations. This practice is sometimes referred to as ``Wiki-Lawyering'' within the community and can signal an unwillingness to compromise.

\subsection{Neural network models}

The results from our neural models illustrate that incorporating structure improves predictive accuracy on our task. 

The model that averages over word embeddings performs the worst; a moderate increase in performance (2\%) is gained from adding sequential word processing by way of an LSTM model. Adding utterance boundary information with HAN  results in a much larger improvement (11\%) over the LSTM. This indicates that, while it is helpful to observe the ordering of words in a conversation, a more critical component in the case of disagreements is how words are arranged in utterances. 

The highest scoring model, which a PR-AUC of 0.40, results from including edit summaries in the conversations ($P<0.05$, using a randomized permutation test). This provides support for our observation that understanding some conversations requires knowledge of the edits that occurred during the conversation. However, neglecting to insert the ``EDIT'' token (as explained in Section \ref{sec:models}) results in the PR-AUC decreasing to 0.33, which indicates that utterances sourced from edit summaries require special processing and are not completely homogeneous in the conversation. 

Evaluating the predictions of this model, we observe that false positives often co-occur with heated debate. This is not unexpected, given that the positive class is contains interlocutors who feel strongly enough about their case to request mediation. The false negatives, on the other hand, contain a number of cases where it seemed as though a disagreement had been resolved, but then the argument progresses as further edits are made. 

\subsection{Early estimation of outcome}
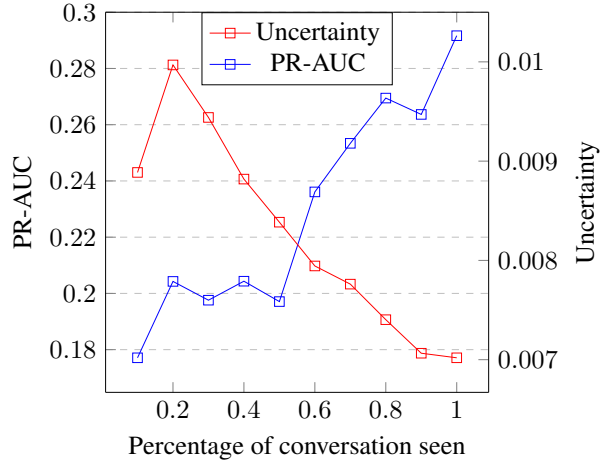
\begin{figure}
\centering
\scalebox{0.9}{
\begin{tikzpicture}
  \pgfplotsset{scale only axis,}
  \begin{axis}[
    ymax=0.3,width=0.35\textwidth,height=0.35\textwidth,
    ymajorgrids=true,
    grid style=dashed,
    axis y line*=left,
    xlabel={Percentage of conversation seen},
    ylabel={PR-AUC},
    legend style={at={(0.5,1)},anchor=north}]
    \addplot[color=blue,mark=square]
            coordinates {
            (0.1,0.177104)(0.2,0.204266)(0.3,0.197588)(0.4,0.204341)(0.5,0.197096)
            (0.6,0.236099)(0.7,0.253375)(0.8,0.269468)(0.9,0.26365)(1,0.291656)
            }; \label{plot_1_y1}
    \end{axis}
    \begin{axis}[
      ymax = 0.0105, width=0.35\textwidth,height=0.35\textwidth,
      axis y line*=right,axis x line=none,
      ylabel={Uncertainty},
      yticklabel style={
      /pgf/number format/fixed,
      /pgf/number format/precision=5},
      scaled y ticks=false,
      legend style={at={(0.5,1)},anchor=north}]
    \addplot[color=red,mark=square]
            coordinates {
            (0.1,0.008886)(0.2,0.009969)(0.3,0.00944)(0.4,0.008819)(0.5,0.008385)
            (0.6,0.007944)(0.7,0.007761)(0.8,0.007403)(0.9,0.007065)(1,0.007019)
            }; \label{plot_1_y2}
    \addlegendimage{/pgfplots/refstyle=plot_1_y1}\addlegendentry{Uncertainty}
    \addlegendimage{/pgfplots/refstyle=plot_1_y2}\addlegendentry{PR-AUC}
  \end{axis}
\end{tikzpicture}}
\caption{PR-AUC scores and uncertainty values of a HAN model when exposed to partial conversations.}
\label{fig:early_estimation}
\end{figure}
\begin{figure*}[]
    \centering
    \includegraphics[width=\textwidth]{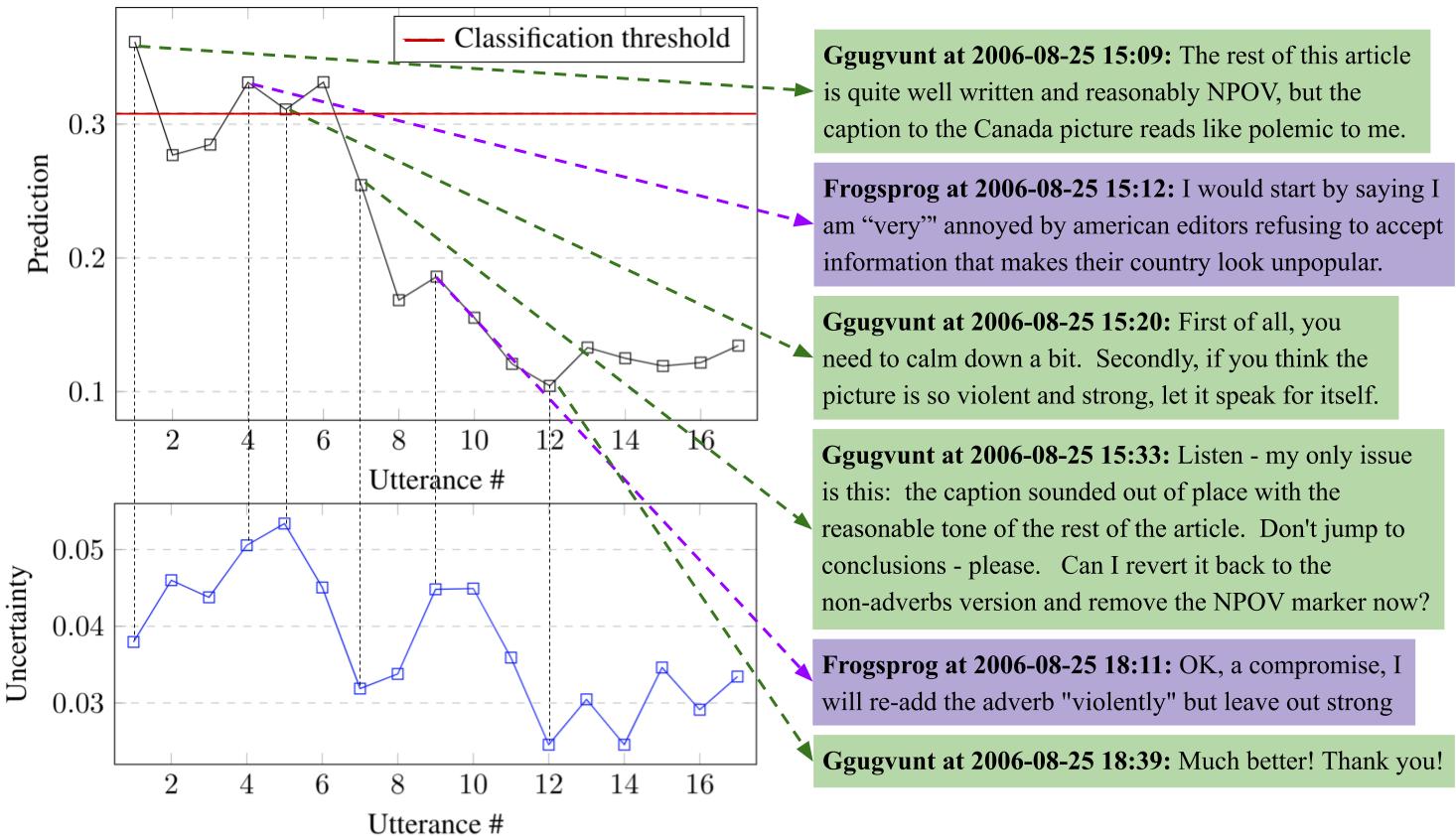}
    \caption{Model predictions and uncertainty throughout the course of a conversation in our dataset.  The classification threshold shown is tuned on the validation set.}
    \label{fig:uncertainty_example}
\end{figure*}

We are interested in how model predictions change throughout the conversation; whether it is possible to predict the outcome from the beginning, or if there is often some inflection point in a conversation that changes the outcome. Predicting the outcome of a conversation based on only a few utterances means that the model has less information for its inference; however, if models can perform well under such conditions, early intervention by a mediator could be recommended.

To evaluate model performance in such conditions, we split each disagreement into 10 buckets, chronologically (using only conversations of more than 10 utterances). We evaluate models trained on full conversations on these subsets to observe the change in model performance. These results are shown in Figure~\ref{fig:early_estimation} in blue. Model performance increases from 0.198 at the midway point (0.5 bucket) to 0.292 when the full conversation has been observed, indicating that signals later in conversations are often important for predicting outcome. However, the neural model's performance when predicting on half the conversation only is still better than the toxicity and sentiment models on the complete conversations.

\subsection{Uncertainty estimation}\label{sec:uncertainty}

Given that this early estimation exposes the model to less information, we are interested in whether this impacts model uncertainty. We employ Monte Carlo dropout \citep{gal2016dropout} to this end. This approach calls for applying dropout before every layer of weights during both training and prediction, allowing us to sample from the distribution of predicted values for each input. Each input is evaluated multiple times (in our case, $N=30$) and the mean and standard deviation of these predictions represent the prediction and its uncertainty. 

Our results are shown in Figure~\ref{fig:early_estimation} (in red). We note that uncertainty initially increases and then decreases monotonically. The initial increase in uncertainty could be due to a contrary position being introduced in the second utterance, in response to the introductory comment (as occurs in Figure~\ref{fig:uncertainty_example}). From there, the dispute is either resolved or eventually escalated, and uncertainty decreases almost linearly as the model is exposed to new data.

\subsection{Identifying inflection points}

Having ascertained that model accuracy improves and uncertainty decreases as a model is exposed to more information, we are interested in factors that cause the predicted class to change, and how the model predictions relate to the feature-based model coefficients. Although methods for interpreting neural network predictions exist \citep{ribeiro2016should,ribeiro2018anchors}, these are not easily extendable to process inter-utterance dependencies as observed with the HAN. We instead analyse a conversation from our dataset to observe inflection points and what may have caused them. We show HAN model predictions and uncertainty values in Figure~\ref{fig:uncertainty_example}. 

The HAN model initially predicts escalation, with a relatively low uncertainty value. The conversation remains confrontational for the first seven utterances with a high escalation score. There are two ``I'' pronouns in the second utterance and two ``you'' pronouns in the third utterance, which were also associated with unconstructiveness in the feature-based models. We note the use of politeness cues (``please'' and ``thank you'' in the fourth and sixth utterances in this example), which reduces the score for escalation. The uncertainty and prediction values increase when one user proposes a compromise in utterance 9, and then decreases again in utterance 12 when the other user accepts the compromise. This feature was not explicitly modelled for in our feature-based models.

\section{Conclusion}

In a discussion of online disagreements by \citet{graham} (which Wikipedia references in its guidelines for resolving disputes), the author remarks that an increase in disagreements through widespread online interaction has the potential to create a surge in anger among internet users. On the other hand, as illustrated in our data, disagreements can sometimes be constructive. The success of Wikipedia shows the benefits of integrating multiple perspectives. As \citet{hahn2020argument} states, ``Arguing things through is at the center of our attempts to come to accurate beliefs about the world [and] decide on a best course of action.'' For this reason, it is important to understand why disagreement can sometimes be constructive, and in other cases leads to conversational failure.

In this work, we investigated constructive disagreements from an NLP perspective. We have proposed a dataset of disagreements online and defined the task of predicting escalation as a proxy label for cases where disagreements were unconstructive. We analysed features that are associated with either class, and drew parallels with existing work. Using neural network models, we investigated the effect of modeling conversation structure and found that adding utterance hierarchy lead to an increase in performance. Finally, we validated our neural models by evaluating their performance and uncertainty when exposed to partial conversations. Our insights on constructive disagreements are not limited to Wikipedia and would be transferable to disagreements on other platforms. Additionally, our finding that edit summaries are an informative part of Talk page conversations should be useful for researchers who work on Wikipedia Talk pages more generally. 

\section*{Acknowledgements}
Christine De Kock is supported by scholarships from Huawei and the Oppenheimer Memorial Trust. Andreas Vlachos is supported the EPSRC grant no. EP/T023414/1: Opening Up Minds.

\bibliography{eacl2021}
\bibliographystyle{acl_natbib}

\appendix
\section{Words list}\label{app:words}
Words associated with the positive and negative class are shown in Table \ref{tab:words}. Coefficients are determined by training a bag-of-words model with logistic regression. The positive class is escalation. 

\begin{table}[]
    \centering
    \begin{tabular}{|l|r|}
    \hline
    \textbf{Words} &   \textbf{Coefficients} \\
    \hline
    npov        & -8.421 \\
    information &  6.138 \\
    did         &  5.583 \\
    wp          &  4.969 \\
    right       & -4.819 \\
    agree       & -4.124 \\
    discussion  & -4.106 \\
    issue       &  3.864 \\
    say         &  3.771 \\
    pov         & -3.754 \\
    want        &  3.610 \\
    article     & -3.398 \\
    work        &  3.362 \\
    articles    &  2.755 \\
    edit        &  2.747 \\
    claim       & -2.232 \\
    case        &  2.180 \\
    people      & -2.140 \\
    wikipedia   &  2.132 \\
    use         & -2.013 \\
    \hline
    \end{tabular}
    \caption{Words associated with the positive and negative class, using a bag-of-words model to predict escalation.}
    \label{tab:words}
\end{table}

\end{document}